%% file: STACOM2026_mainpaper.tex
\documentclass[runningheads]{llncs}
\input{settings/packages}
\input{settings/commandes_math}

\input{settings/custom}
\usepackage[T1]{fontenc}

\usepackage{graphicx,verbatim}
\usepackage[normalem]{ulem}

\usepackage{color}

\begin{document}
\title{TSPFN: A Temporal Tabular Foundation Model for Physiological Time Series Classification}

\author{Jérémie Stym-Popper\inst{1} \and Clément Rambour\inst{1}\and Federica Granese\inst{2}\and Nicolas Thome\inst{1,5} \and Olivier Bernard\inst{4,5}}  
\authorrunning{J. Stym-Popper et al.}
\institute{Sorbonne Université, CNRS, ISIR, MLIA, F-75005 Paris, France\and Inria, CNRS, I3S, Université Côte d’Azur, Valbonne, France \and INSA-Lyon, Université Claude Bernard Lyon 1, CNRS, Inserm, CREATIS UMR 5220, U1294,
F-69621, LYON, France \and Institut Universitaire de France (IUF)\\
    \email{stympopper@isir.upmc.fr}}
  
\maketitle              
\begin{abstract}

Designing models that generalize effectively in low- to medium-data regimes remains a primary challenge in medical machine learning, particularly for physiological time-series classification. While tabular foundation models such as TabPFN offer an attractive alternative to conventional fine-tuning through in-context learning, they are not designed to capture the temporal dependencies inherent to physiological signals.
~In this paper, we introduce TSPFN, a foundation model that redesigns TabPFN's architecture for time series data. TSPFN integrates structured temporal representations and positional embeddings to capture intra-sample temporal and channel dependencies. To fully leverage its spatio-temporal design, the model is pretrained on 140,000 real-world physiological time series across multiple medical domains. This yields a unified, generalizable framework capable of learning the specificities of medical time series. Experiments across diverse physiological benchmarks demonstrate that TSPFN consistently outperforms standard tabular baselines and TabPFN, and achieves superior cross-domain generalization compared to specialized deep time-series models. All our experiments, ablation studies, and pre-processing scheme are publicly available at \texttt{\url{https://github.com/Jeremstym/TSPFN}}


\end{abstract}

\section{Introduction}
\label{sec:introduction}
Physiological time series play a central role in many clinical workflows, serving as compact and informative representations derived from raw medical signals or images. However, learning robust predictive models for classification tasks from such data remains challenging, particularly in low- to medium-data regimes that are common in clinical practice. The strong inter-subject variability, heterogeneous temporal dynamics, and limited availability of labeled recordings substantially hinder the generalization ability of conventional supervised learning approaches, which typically rely on task-specific training and large annotated datasets. As a result, models trained on small or moderately sized cohorts often fail to generalize across datasets, modalities, or clinical settings. 

Tabular foundation models (TFMs) have recently emerged as a powerful alternative to classical machine learning approaches in low-data regimes. A TFM is a neural architecture pretrained on large corpora of diverse tabular datasets, enabling it to perform a wide range of downstream supervised tasks through in-context learning without gradient-based fine-tuning.
A prominent example is TabPFN~\cite{hollmann2023tabpfn}, a pretrained transformer that learns to approximate Bayesian inference on synthetic datasets sampled from a prior defined over a large space of structural causal models, enabling zero-shot generalization to new tasks at test time.
While TabPFN is highly effective in classical tabular settings, its reliance on permutation-invariant features and synthetic pretraining datasets limits its applicability to medical time series, where intrinsic temporal dependencies and inter-channel structure are central.   

In parallel, recent works have explored extending TabPFN to time-series forecasting~\cite{cai2025explore,hoo2024tabular,hoo2025tables}. Rather than modeling time series as sequential signals, these approaches reformulate forecasting as a tabular regression problem, transforming each time series into a feature vector encoding temporal information via cyclic trigonometric representations of calendar attributes. 
While this reformulation yields strong performance on general-purpose benchmarks, it relies on the assumption that temporal structure can be adequately captured through calendar-based features, an assumption often violated in the medical domain where physiological signals exhibit complex intrinsic temporal dependencies. Crucially, this formulation is inherently designed for forecasting future time points and does not naturally extend to time series classification tasks, which require reasoning over entire temporal trajectories.

To address these limitations, we introduce \textbf{TSPFN} (Time-Series Prior-Data Fitted Network), a novel
~in-context learning model tailored for the classification of physiological time series datasets in small- and medium-scale data regimes.
Our main contributions are summarized as follows:
\begin{itemize}
    \item From a methodological perspective, we identify the main limitations of classical tabular foundation models in handling time series physiological data, and we address them by proposing TSPFN, which
    revisits the 
    TabPFN architecture 
    to model temporal dependencies in multivariate, multi-channel time series representations by integrating structured input representations and channel-wise positional embeddings to capture intra-sample temporal dependencies in physiological time series (cf.~\Cref{sec:method}).
    
    \item From an experimental perspective, we build a large-scale collection of real-world physiological time series from publicly available clinical datasets
    encompassing EEGs, ECGs, and ICU waveforms ($\sim$140,000 samples). Moreover,
    we organize them into a unified multi-scale pretraining framework with a consistent preprocessing scheme, which we release publicly\footnote{\url{https://github.com/Jeremstym/TSPFN}.} (cf.~\Cref{sec:experiments}).
    \item From an evaluation perspective, we benchmark TSPFN on five diverse physiological time series datasets, showing competitive or superior performance against state-of-the-art methods in the low- to medium-data regimes, and excellent cross-domain generalization performances.
    ~We further conduct an ablation study to assess the contributions of the proposed temporal representations and positional encoding strategies to overall model performance (cf.~\Cref{sec:results}). 
\end{itemize}

\section{Background}

The proposed model builds upon Prior-Data Fitted Networks (PFNs)~\cite{mullertransformers}, a class of neural networks approximating Bayesian prediction in an in-context learning setting, i.e., without parameter updates at test time. PFNs are transformer-based models pre-trained on synthetic datasets sampled from a prior distribution over tasks. Through this process, the model learns to map a set of labeled examples (the context) to predictive distributions for unseen query samples.
During pretraining, each sampled dataset is split into a support set and a query set, and the model is optimized to predict the query labels given the support examples. At inference time, the trained PFN can be directly applied to new downstream tasks without fine-tuning, provided that the target data distribution is consistent with the prior at pretraining. Given a small set of labeled examples, the model produces predictions for new inputs in a single forward pass. 

\section{Method: TSPFN}
\label{sec:method}
\begin{figure}[t]
    \centering
    \includegraphics[width=\linewidth]{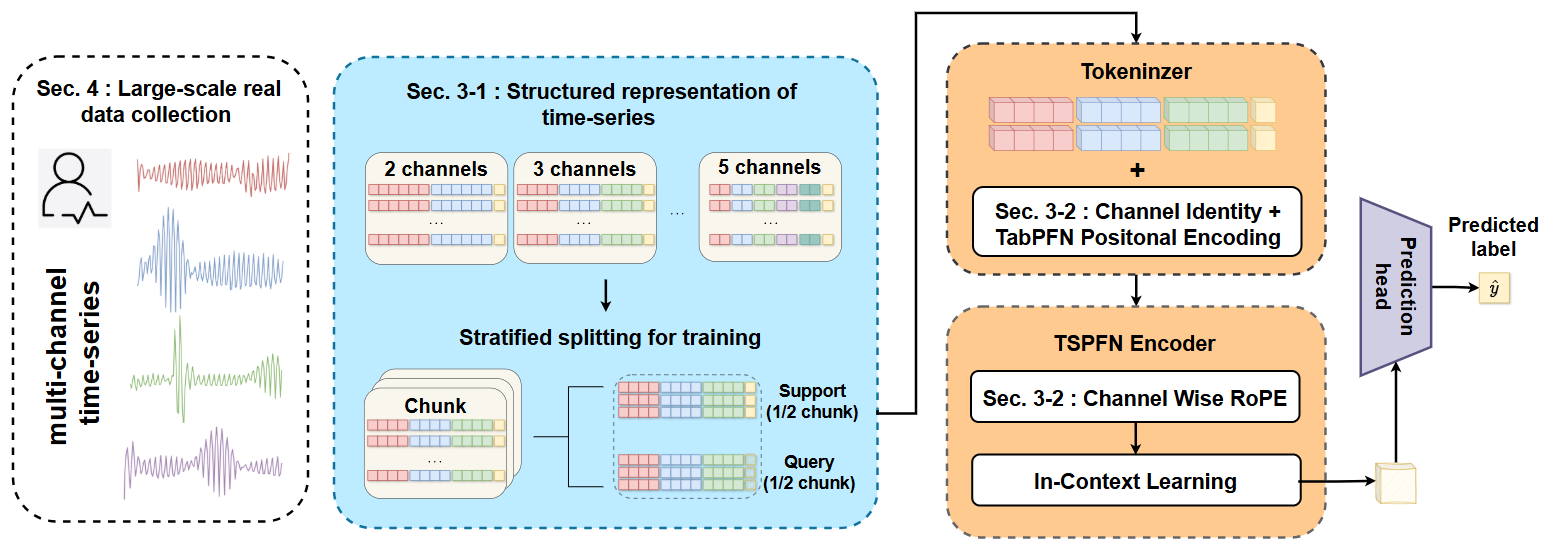}
    \caption{Overview of the proposed TSPFN framework. Physiological time series are structured into a unified tabular representation with explicit temporal and channel organization, enabling in-context learning for time series classification.}
    \label{fig:dataset-preprocessing}
\end{figure}
TabPFN is a pretrained transformer that learns to approximate Bayesian inference on synthetic datasets sampled from a prior defined over a large space of structural causal models (cf.~\Cref{sec:introduction}). However, generating realistic synthetic physiological time series with associated labels is inherently difficult, making TabPFN's synthetic data generation strategy unsuitable in this domain. To address this limitation, 
we leverage the growing availability of open-access physiological data and construct a large-scale collection of nearly 140{,}000 real-world time series spanning multiple modalities, including electrocardiography (ECG), electroencephalography (EEG), and vital signals. More details on the real-world corpus underlying TSPFN are provided in \Cref{sec:experiments}.

\subsection{Structured representation of physiological time series}

While the original TabPFN operates on unordered tabular features, our approach restructures the input space to explicitly reflect the temporal and channel-wise organization of physiological data. As illustrated in~\Cref{fig:dataset-preprocessing}, each row of the input tabular data corresponds to one or more patient-specific physiological time series, followed by the ground-truth label.
Since TSPFN builds upon the TabPFN architecture, each input row is constrained to a maximum feature dimensionality of $F_{\max}=500$. To accommodate variability in sequence length and channel count across datasets, we adopt a multi-scale representation strategy by varying both the sequence length $T$ and the number of channels $C$ during training, while enforcing the constraint $T \times C \leq F_{\max}$. This encourages the model to learn scale-invariant temporal representations. Concretely, we consider the following configurations: $(T, C) \in \{(250,2), (166,3), (125,4), (100,5)\}$.
Each input row is formed by concatenating channel-wise signals under this constraint, yielding a structured representation that preserves both temporal and channel organization, as illustrated in~\Cref{fig:dataset-preprocessing}.

\subsection{Structured temporal and channel embeddings}
\begin{figure}[t]
    \centering
    \includegraphics[width=0.99\linewidth]{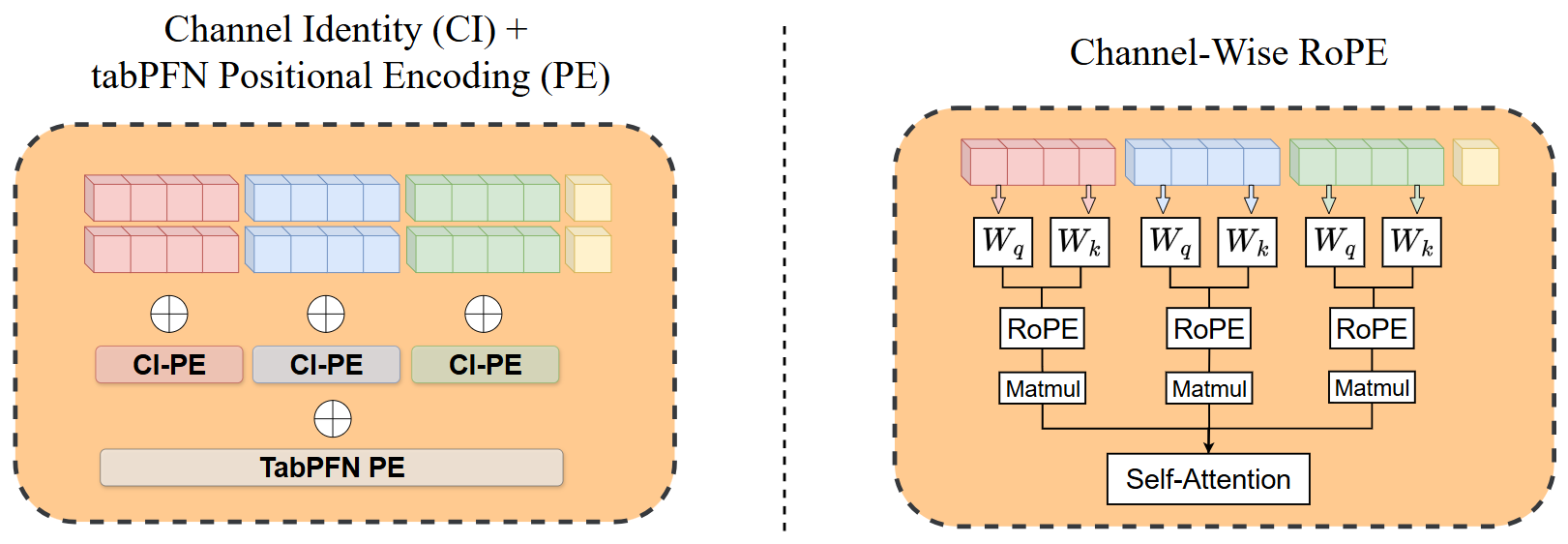}
    \caption{\textbf{Left} Cross-channel dependencies are modeled by adding the positional encoding from~\cite{hollmann2023tabpfn} (PFN-PE) to the concatenated multi-channel input while each channel is identified by a learnable channel embedding (CI-PE). \textbf{Right} Intra-channel dependencies are captured via multiplication of the query and key of each channel by the RoPE matrix before self-attention.   }
    \label{fig:PE}
\end{figure}

To adapt TabPFN to physiological time series, we introduce additional positional encodings that explicitly model temporal and channel structure
\vspace{0.2cm}~\\
\textbf{Channel-Wise Rotary Temporal Embeddings.}
To handle variable-length time series while preserving order, we adopt Rotary Position Embeddings (RoPE)~\cite{SU2024127063}. Following~\cite{li2025mira}, we employ a channel-wise variant in which temporal positional information is injected independently for each channel by applying position-dependent rotations to the query and key representations in the attention mechanism, as illustrated in~\Cref{fig:PE} (b). For a time step $t$, the rotation angle is $\theta_j(t) = t \cdot 10000^{-2j/d}$, where $j \in \{0, \dots, \frac{d}{2}-1\}$ indexes the feature dimensions and $d$ denotes the hidden dimensionality.
The rotary positional encoding depends only on the temporal index and is shared across channels. This ensures that attention depends on relative temporal offsets rather than absolute positions, enabling consistent modeling of temporal dependencies across variable-length sequences.
\vspace{0.2cm}~\\
\textbf{Channel Identity Embeddings.}
To preserve channel identity within the concatenated input sequence, we introduce learnable channel embeddings as can be seen in~\Cref{fig:PE} (a). A lookup table $\mathcal{C} \in \mathbb{R}^{C \times d}$ assigns a channel-specific embedding to each input, which is added to the corresponding representation and shared across all time steps of that channel. This allows the model to capture inter-channel dependencies while jointly modeling temporal dynamics.

\subsection{Training procedure}
During pretraining, datasets are partitioned into chunks of $N=5{,}000$ samples. Each chunk is split into a balanced support set $\mathcal{S}$ and query set $\mathcal{Q}$ of equal size. Samples in $\mathcal{S}$ include both time series inputs and ground-truth labels, while labels in $\mathcal{Q}$ are replaced by a learnable mask symbol. Stratified splitting ensures identical class distributions across $\mathcal{S}$ and $\mathcal{Q}$. Predictions corresponding to masked labels in $\mathcal{Q}$ are processed by a projection head, and the model is trained end-to-end using a cross-entropy loss computed exclusively on the query set.

\section{Experiments}
\label{sec:experiments}

\subsubsection{Experimental setup.}
TSPFN is initialized from pretrained TabPFN weights and optimized using the procedure described above. Optimization is performed with AdamW on a single NVIDIA H200 GPU, using a fixed learning rate of $5\times10^{-5}$, weight decay $0.1$, $\epsilon = 10^{-8}$, and $\beta = (0.9, 0.999)$. The model is trained for up to 100 epochs (10{,}000 steps), with early stopping based on the validation cross-entropy loss; in practice, convergence is typically reached after  $\sim$25 epochs.\vspace{0.2cm}~\\ 
\textbf{Datasets.} (\textit{Pretraining})
To boost generalization, we build a pretraining corpus spanning multiple physiological domains with heterogeneous temporal dynamics, including EEG, ECG, and ICU waveforms. We leverage four publicly available datasets: \textbf{TUEV}~\cite{He_2020_CVPR} and \textbf{TUAB}~\cite{7405423} for EEG, \textbf{PTB-XL}~\cite{wagnerPTBXLLargePublicly2020} for ECG, and \textbf{HiRID}~\cite{hylandEarlyPredictionCirculatory2020} for multivariate ICU signals, yielding a total of $\sim$140,000 samples. To focus pretraining on local temporal dependencies, we restrict sequence lengths to 100–250 time points. Robustness to varying channel dimensions is promoted by representing each sample under multiple channel configurations ($C \in \{2,3,4,5\}$).

(\textit{Evaluation}) We evaluate TSPFN under realistic clinical constraints on a diverse physiological benchmark characterized by limited data, class imbalance, and high intra-class variability.
All experiments use stratified 5-fold cross-validation. The benchmark includes five datasets: \textbf{eICU-CRD}~\cite{pollardEICUCollaborativeResearch2018} for ICU, \textbf{ESR}~\cite{Andrzejak2001IndicationsON}, \textbf{EOS}~\cite{Bagnall2018TheUM}, for EEG, \textbf{ECG5000}~\cite{Chen2015AGF}, and \textbf{CPSC 2018}~\cite{Ng2018AnOA} for ECG. Results are reported as the mean performance across folds.\vspace{0.1cm}~\\ 

\subsubsection{Pretraining Datasets.}
To enhance the model’s generalization capabilities, we curate a pretraining database across four distinct physiological domains. This corpus encompasses a broad spectrum of temporal dynamics, ranging from high-resolution electrocardiograms (ECG) and neuro-oscillatory electroencephalograms (EEG) to lower-frequency intensive care unit (ICU) waveforms. By pretraining on these diverse morphologies, the model learns robust representations of varied physiological patterns. We utilize the following datasets: (i) \textbf{TUEV} \cite{He_2020_CVPR}: An EEG corpus annotated for six event types (e.g., spikes, sharp waves, and background activity). We extract 8,993 training and 2,865 validation windows. (ii) \textbf{TUAB} \cite{7405423}: A large-scale EEG dataset labeled for normal vs. abnormal clinical findings, comprising 32,415 training and 8,113 validation windows. (iii) \textbf{PTB-XL }\cite{wagnerPTBXLLargePublicly2020}: A 12-lead ECG dataset. We formulate a binary pretext task (normal vs. abnormal) using a subset of up to five leads. Lead II serves as a temporal reference for R-peak detection and windowing, resulting in 48,335 training and 5,483 validation samples. (iv) \textbf{HiRID} \cite{hylandEarlyPredictionCirculatory2020}: A dataset of high-acuity ICU variables. We select five specific measurements: heart rate, dobutamine dosage, mean arterial pressure (MAP), oxygen saturation (SpO2), and peak inspiratory pressure. To prevent data leakage and focus the pretraining on capturing local temporal dependencies, we restrict sequences to a range of 100 to 250 time points (approximately 8.3 to 20.8 hours). This setup yields 26,879 training and 6,719 validation sequences focused on local signal dynamics. To facilitate  multi-channel robustness, each sample is represented four times in our database with varying channel depths ($C \in \{2, 3, 4, 5\}$).

\subsubsection{Evaluation Datasets} 

To evaluate the generalization of TSPFN under real-world clinical constraints, we construct a benchmark spanning three primary physiological modalities—vital signs, EEG, and ECG. These tasks reflect low-data regimes, high intra-class variability, and heavy class imbalance. Evaluated via stratified 5-fold cross-validation on subsampled sets of $N=100$ per task (Table \ref{100samples-bench}), all inputs undergo standard domain-specific preprocessing as detailed below. \textbf{eICU-CRD} \cite{pollardEICUCollaborativeResearch2018} is a multivariate vital-sign dataset using 5 continuous channels (heart rate, respiration rate, $\text{SpO}_2$, mean arterial blood pressure, and body temperature). To establish a non-trivial prediction horizon for ICU mortality/discharge outcome, we extract a 100-point observation window ($\approx 8.3$ hours at a 5-minute sampling interval) ending exactly 4 hours prior to patient discharge. Epileptic Seizure Recognition \textbf{ESR} \cite{Andrzejak2001IndicationsON} comprises 11,500 single-channel EEG segments, each containing 178 time points ($\approx 1$ second duration sampled at 173.61 Hz). The task evaluates multi-class classification across 5 balanced clinical states ranging from seizure activity to healthy control conditions. EyesOpenShut (\textbf{EOS}) \cite{Bagnall2018TheUM} is a small-scale, binary classification EEG dataset from the UEA Multivariate Time Series Archive ($N=98$ recordings, each of length 128). To guarantee stability across baselines and avoid noise artifacts, we extract a 3-channel subset corresponding to electrodes T7, P7, and F4 (channels 4, 5, and 11). \textbf{ECG5000} \cite{Chen2015AGF} consists of 5,000 single-lead, high-resolution ECG heartbeats (140 time points per length) from the UCR archive. Rather than collapsing the dataset into a standard binary (normal vs. abnormal) setup, we preserve the original, highly imbalanced 5-class distribution to maintain diagnostic complexity. \textbf{CPSC} 2018 \cite{Ng2018AnOA} contains variable-length 12-lead ECG recordings. We filter for sequences $\ge 10$ seconds, select the first 4 leads, and map the original 9 arrhythmia categories into 4 consolidated, clinically interpretable diagnostic groups. Following the PTB-XL windowing protocol, we extract 125-point sub-sequences centered on detected R-peaks, yielding 10,636 standardized windows for classification.

\section{Results and ablation studies}
\label{sec:results}

\subsubsection{Baselines and SOTA.}
Our experimental results demonstrate that TSPFN consistently performs at or above the current state-of-the-art (SOTA). To rigorously evaluate the versatility of our approach, we benchmark it against five distinct baselines, each representing a fundamentally different paradigm in time series classification—ranging from traditional statistical boosting to modern foundation models. (1) \textbf{XGBoost}~\cite{10.1145/2939672.2939785}, a dominant tree-based model optimized here for high-dimensional generalization via shallow trees (max depth: 3), aggressive temporal subsampling (colsample bytree: 0.3), and conservative learning rates to mitigate signal noise; (2) \textbf{TabPFN}~\cite{hollmann2023tabpfn}, a 7.3M-parameter competitive in-context transformer for tabular tasks. By extending the TabPFN-v2 framework to the time series domain at a matching scale of 7.3M parameters, TSPFN directly highlights the impact of our architectural adaptations; (3) \textbf{Temporal Convolution Network (TCN)}~\cite{BaiTCN2018}, a 356K-parameter deep convolutional SOTA baseline for time series, utilizing dilated causal layers to capture long-range temporal dependencies and morphological pattern recognition in ECG signals. We utilize a residual causal architecture with 64 filters ($F_t$), a kernel size of 11 ($K_t$), and 20\% dropout ($P_t$), feeding the last-step representation into the prediction head for final classification; (4) \textbf{MiniRocket}~\cite{dempster_etal_2021}, a lightweight, deterministic kernel-based baseline featuring a parameter-free feature extractor paired with a low-capacity linear head ($\approx$100K trainable parameters for a 10-class task), providing a highly efficient benchmark via the Proportion of Positive Values (PPV) transform; and (5) \textbf{LaBraM}~\cite{jiang2024large}, a 7.5M-parameter physiological foundation model. Utilizing a neural tokenizer and massive pre-training on EEG signals, it benchmarks TSPFN against the cutting edge of transfer learning and cross-domain signal understanding. By leveraging a transformer-based architecture and multi-domain pre-training, TSPFN consistently outperforms established gold-standard baselines such as MiniRocket and LaBraM, demonstrating superior adaptability to varied physiological signals.
\begin{table}[t]
\centering
\footnotesize 
\setlength{\tabcolsep}{1pt} 
\renewcommand{\arraystretch}{0.9} 
\caption{Comparison of state-of-the-art time series classification methods with TSPFN. The best and second-best results are highlighted in blue. Results are computed using 100 support samples. Methods are grouped into in-context learners (TSPFN, TabPFN), fitted tree-based methods ($\dagger$), and fine-tuned deep learning models ($*$).}\label{100samples-bench}

\begin{threeparttable}
\begin{tabular}{cc|ccccc|c}
\toprule
Data& Metric & TabPFN & XGBoost$^{\dagger}$ & TCN$^{*}$ & MiniRocket$^{*}$ & LaBraM$^{*}$ & \textbf{TSPFN} \\
\midrule
\multirow{4}{*}{\rot{Average}} 
 & AUC  &$77.5 \pm 9.7$&$77.1 \pm 9.1$&$74.8 \pm 12.7$&\cellcolor{cyan!15}$78.1 \pm 11.5$&$64.8 \pm 12.5$&\cellcolor{cyan!15}\bm{$82.5 \pm 5.5$}\\
 & PRC  &$49.2 \pm 15.8$&$51.0 \pm 12.8$&$49.5 \pm 10.1$&\cellcolor{cyan!15}$53.6 \pm 11.2$&$37.3 \pm 13.1$& \cellcolor{cyan!15}\bm{$57.3 \pm 11.9$}\\
 & F1   &$42.7 \pm 12.6$&\cellcolor{cyan!15}$44.3 \pm 7.7$&$40.8 \pm 17.9$&$38.4 \pm 15.0$&$13.3 \pm 15.0$&\cellcolor{cyan!15}\bm{$53.3 \pm 7.8$}\\
 & Recall   &\cellcolor{cyan!15}$41.9 \pm 13.0$&$41.7 \pm 8.4$&$41.6 \pm 19.2$&$37.9 \pm 16.7$& $18.6 \pm 15.5$&\cellcolor{cyan!15}\bm{$55.4 \pm 9.8$}\\
\midrule[1.5pt]
\multirow{4}{*}{\rot{eICU}} 
 & AUC  &$66.4 \pm 8.7$&$67.9 \pm 5.8$& $76.9 \pm 3.0$ & \cellcolor{cyan!15}\bm{$80.0 \pm 3.4$} &$62.8 \pm 6.8$& \cellcolor{cyan!15}$79.4 \pm 1.7$\\
 & PRC  &$28.6 \pm 8.1 $&$39.1 \pm 6.3$& \cellcolor{cyan!15}$48.2 \pm 4.9$ & \cellcolor{cyan!15}\bm{$52.0 \pm 8.7$} &$34.0 \pm 7.0$& $48.1 \pm 3.9$\\
 & F1   &$22.3 \pm 14.3 $&$34.4 \pm 8.6$& \cellcolor{cyan!15}$40.1 \pm 3.7$ & $36.3 \pm 13.2$ &$0.0 \pm 0.0$& \cellcolor{cyan!15}\bm{$49.8 \pm 3.4$}\\
 & Recall   &$19.1 \pm 13.8$&$30.1 \pm 12.9$& \cellcolor{cyan!15}$31.2 \pm 4.4$ & $27.5 \pm 14.9$ &$0.0 \pm 0.0$& \cellcolor{cyan!15}\bm{$54.4 \pm 7.3$}\\
\midrule
\multirow{4}{*}{\rot{ESR}} 
 & AUC  &$70.7 \pm 2.1$&$66.5 \pm 1.0$&$81.8 \pm 2.4$ & \cellcolor{cyan!15}\bm{$89.8 \pm 0.6$} &$52.9 \pm 3.7$& \cellcolor{cyan!15}$82.7 \pm 1.6$\\
 & PRC  &$41.5 \pm 2.2$&$38.3 \pm 1.6$&$54.0 \pm 5.1$ & \cellcolor{cyan!15}\bm{$68.3 \pm 1.7$} &$22.5 \pm 2.0$& \cellcolor{cyan!15}$57.7 \pm 2.3$\\
 & F1   &$40.4 \pm 1.9$&$37.4 \pm 1.5$&$49.4 \pm 5.7$ & \cellcolor{cyan!15}\bm{$57.1 \pm 5.3$} &$6.8 \pm 0.4$& \cellcolor{cyan!15}$55.4 \pm 1.8$\\
 & Recall &$41.0 \pm 1.7$&$37.3 \pm 1.2$&$48.5 \pm 5.4$ & \cellcolor{cyan!15}\bm{$60.9 \pm 3.2$} &$20.0 \pm 0.1$& \cellcolor{cyan!15}$56.8 \pm 1.5$\\
\midrule
\multirow{4}{*}{\rot{EOS}} 
 & AUC  & \cellcolor{cyan!15}$77.9 \pm 6.9$&$75.0 \pm 6.8$&$54.9 \pm 11.0$&$58.9 \pm 5.0$ &$47.2 \pm 7.4$& \cellcolor{cyan!15}\bm{$82.8 \pm 2.3$}\\
 & PRC  & \cellcolor{cyan!15}$80.4 \pm 4.9$&$78.5 \pm 6.7$&$59.4 \pm 14.2$&$58.4 \pm 7.0$ &$52.2 \pm 6.0$& \cellcolor{cyan!15}\bm{$85.2 \pm 3.0$}\\
 & F1   & \cellcolor{cyan!15}$63.8 \pm 2.6$&$60.4 \pm 5.9$&$0.0 \pm 0.0$&$41.1 \pm 36.1$ &$22.2 \pm 38.5$& \cellcolor{cyan!15}\bm{$70.0 \pm 1.6$}\\
 & Recall & \cellcolor{cyan!15}$58.7 \pm 2.8$&$55.6 \pm 5.5$&$0.0 \pm 0.0$&$47.6 \pm 47.6$ &$33.3 \pm 57.7$& \cellcolor{cyan!15}\bm{$76.2 \pm 4.8$}\\
\midrule
\multirow{4}{*}{\rot{ECG5000}} 
 & AUC & \cellcolor{cyan!15}\bm{$91.9 \pm 1.8$} & \cellcolor{cyan!15}$91.4\pm 1.1$&$90.1 \pm 1.3$&$86.8 \pm 4.1$&$87.1 \pm 5.2$& $91.3 \pm 2.2$\\
 & PRC &$53.0 \pm 5.3$&$53.1 \pm 7.3$ & \cellcolor{cyan!15}$54.5 \pm 5.0$&$50.3 \pm 7.8$ &$47.2 \pm 2.5$& \cellcolor{cyan!15}\bm{$56.0 \pm 1.8$}\\
 & F1  &$50.6 \pm 8.6$&$50.4 \pm 7.8$ & \cellcolor{cyan!15}\bm{$52.3 \pm 6.4$}&$41.0 \pm 4.4$ &$40.5 \pm 3.5$& \cellcolor{cyan!15}$52.0 \pm 3.0$\\
 & Recall  &$50.9 \pm 8.2$&$49.1 \pm 6.7$ & \cellcolor{cyan!15}\bm{$51.9 \pm 5.1$}&$41.2 \pm 3.5$ &$41.5 \pm 2.9$& \cellcolor{cyan!15}$51.6 \pm 3.4$\\
\midrule
\multirow{4}{*}{\rot{CPSC}} 
& AUC & \cellcolor{cyan!15}$75.6 \pm 2.3$&$74.6 \pm 2.2$&$69.2 \pm 2.6$&$67.6 \pm 1.8$ &$64.8 \pm 3.2$& \cellcolor{cyan!15}\bm{$76.5 \pm 1.4$} \\
& PRC & \cellcolor{cyan!15}$50.1 \pm 3.2$&$48.1 \pm 3.4$&$43.8 \pm 2.5$&$40.6 \pm 2.5$ &$36.2 \pm 1.8$& \cellcolor{cyan!15}\bm{$50.8 \pm 1.3$} \\
& F1  &$41.3 \pm 2.6$ & $44.4 \pm 3.7$ & \cellcolor{cyan!15}$44.7 \pm 3.0$ & $31.1 \pm 3.0$ &$15.3 \pm 0.1$& \cellcolor{cyan!15}\bm{$45.7 \pm 1.8$} \\
& Recall &$43.3 \pm 2.4$ & \cellcolor{cyan!15}$45.1 \pm 3.3$ & $45.0 \pm 2.6$ & $35.4 \pm 1.9$ &$25.3 \pm 0.7$&\cellcolor{cyan!15}\bm{$46.3 \pm 1.6$}\\
\bottomrule
\end{tabular}
\end{threeparttable}
\end{table}

\textbf{Comparative performance analysis.}
Results in the low-data regime (100 training/support samples) are presented in \Cref{100samples-bench}. The average performance across the five datasets shows that TSPFN significantly outperforms baseline methods such as XGBoost and tabular foundation models (TabPFN), as well as specialized deep time series models like TCN, MiniRocket, and LaBram, which require fine-tuning on each dataset. A DeLong test vs MiniRocket in AUROC shows a p-value of $3.7 \times 10^{-5}$. This clearly demonstrates the relevance of foundation models for time series classification and highlights the effectiveness of in-context learning for physiological signals in data-constrained settings. Unlike gradient-based fine-tuning—which often leads to overfitting in high-capacity models such as LaBraM or TCN—TSPFN leverages small support sets without requiring parameter updates. Finally, the temporal inductive bias introduced during pretraining allows TSPFN to consistently outperform general-purpose baselines.
A closer analysis of individual dataset performance reveals that while specialized baselines may occasionally outperform TSPFN on their dedicated benchmarks, they fail to generalize across other domains. In contrast, TSPFN maintains stable and competitive performance across all physiological datasets. 
\vspace{0.2cm}~\\

\begin{figure}[!h]
    \centering
    \includegraphics[width=\linewidth]{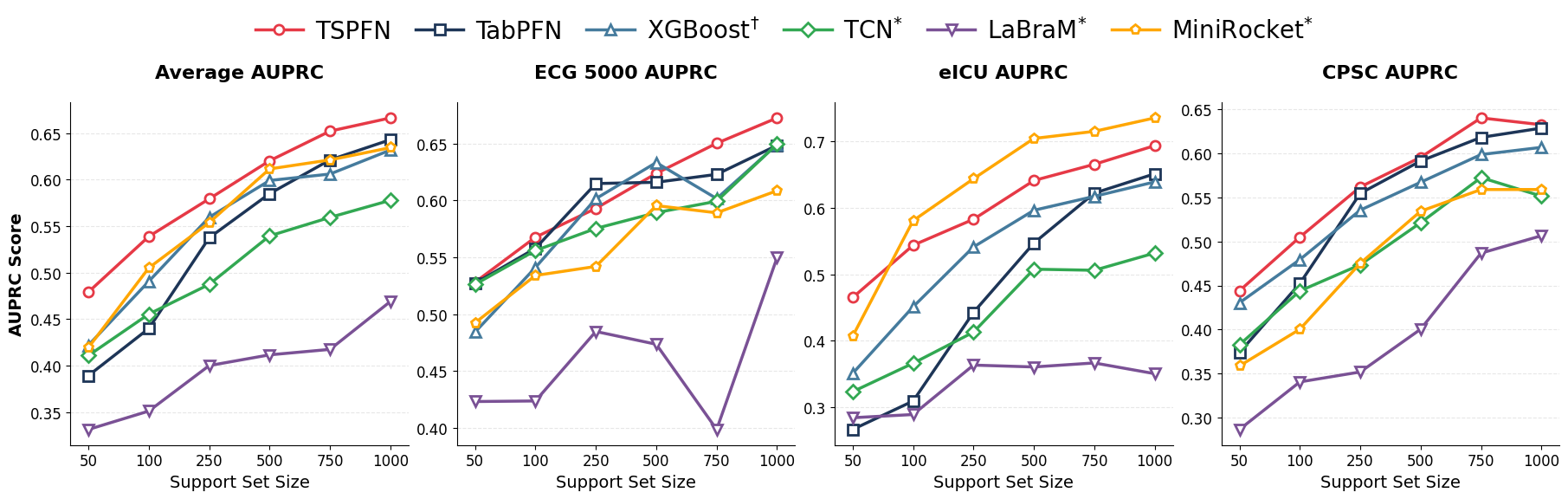}
    \caption{Cross-domain generalization across different support set regimes. Mean AUPRC scores averaged over 3-fold cross-validation for TSPFN and SOTA baselines. Methods are grouped into in-context learners (TSPFN, TabPFN), fitted tree-based methods ($\dagger$), and fine-tuned deep learning models ($*$).}
    \label{fig:histograms}
\end{figure}

\textbf{Cross-domain generalization.}
To further assess the good generalization properties of our model, we scale the data regime from low to medium (up to 1,000 samples). The corresponding results are given in~\Cref{fig:histograms}. TSPFN achieves top-1 AUC on average across all data scales, showing consistent cross-domain performances. 
While some baselines achieve comparable performance to TSPFN on individual datasets, they exhibit substantial variability across the benchmark suite. MiniRocket is specifically tailored to EEG signals and may even outperform TSPFN on eICU; however, its performance degrades markedly on ECG datasets such as CPSC. Finally, EEG-specific foundation models such as LaBraM perform poorly in these settings, with the worst performance across all modalities and data scales. \vspace{0.2cm}~\\ 

\textbf{Ablation studies.}
\Cref{ablations} presents an ablation study assessing the impact of two key components, including broad-spectrum pretraining (TSPFN), and specific channel-wise and time series positional encoding (+RoPE \& CWPE). Pretraining on physiological time series already yields significant gains over TabPFN: +10 points in AUPRC on EOS and +7.8 points on ESR. The proposed temporal encoding further increases these gains, leading to significant improvements across all benchmarks: +6.4 points on ESR and +8.3 points on ECG5000 (coupled with RoPE)
compared to the tabular baseline. Overall, dedicated pretraining and temporal channel-wise encoding greatly enhance the generalization capability of TSPFN compared to TabPFN.

\definecolor{diffgreen}{rgb}{0, 0.6, 0}

\begin{table}[t]
\caption{Ablation study of TSPFN on five validation datasets using the full support set. Performance is reported as AUPRC. The TSPFN column reports results obtained without the proposed positional embeddings, while +(RoPE \& CWPE) corresponds to the complete model incorporating both RoPE and channel-wise positional embeddings.}
\label{ablations}
\centering
\footnotesize
\begin{threeparttable}
\setlength{\tabcolsep}{4pt} 
\begin{tabular}{c|cccc}
\toprule
\textbf{Dataset} & \textbf{TabPFN} & \textbf{TSPFN} & \textbf{+RoPE} & \textbf{+(RoPE \& CWPE)} \\
\midrule
eICU-CRD & 68.3 & 70.5 \textcolor{TealBlue}{\scriptsize($\blacktriangle$2.2)} & 70.6 \textcolor{TealBlue}{\scriptsize($\blacktriangle$0.1)} & 72.9 \textcolor{TealBlue}{\scriptsize($\blacktriangle$2.3)} \\
ESR & 48.8 & 56.6 \textcolor{TealBlue}{\scriptsize($\blacktriangle$7.8)} & 71.8 \textcolor{TealBlue}{\scriptsize($\blacktriangle$15.2)} & 78.2 \textcolor{TealBlue}{\scriptsize($\blacktriangle$6.4)} \\
EOS & 63.3 & 74.0 \textcolor{TealBlue}{\scriptsize($\blacktriangle$10.7)} & 78.4 \textcolor{TealBlue}{\scriptsize($\blacktriangle$4.4)} & 80.7 \textcolor{TealBlue}{\scriptsize($\blacktriangle$2.3)} \\
ECG5000 & 66.1 & 66.7 \textcolor{TealBlue}{\scriptsize($\blacktriangle$0.6)} & 49.1 \textcolor{YellowOrange}{\scriptsize($\blacktriangledown$17.6)} & 74.4 \textcolor{TealBlue}{\scriptsize($\blacktriangle$25.3)} \\
CPSC & 65.1 & 65.1 \textcolor{TealBlue}{\scriptsize($\blacktriangle$0.0)} & 33.2 \textcolor{YellowOrange}{\scriptsize($\blacktriangledown$31.9)} & 65.3 \textcolor{TealBlue}{\scriptsize($\blacktriangle$32.1)} \\
\bottomrule
\end{tabular}
\end{threeparttable}
\end{table}

\section{Conclusion}
We introduced TSPFN, a novel PFN-based model tailored for the classification of physiological time series datasets in the small- and medium-data regime (datasets with fewer than 1,000 samples). The TSPFN architecture builds on TabPFN while addressing three key limitations: permutation-invariant features that ignore temporal dependencies, the lack of position encoding for sequential data, and treating multi-channel signals as independent features. TSPFN overcomes these through a structured multi-scale input representation, channel-wise RoPE for temporal ordering, and channel identity embeddings for inter-channel dependencies. 
Extensive evaluation on EEG, ECG, and ICU datasets shows that TSPFN, despite being pretrained in heterogeneous physiological signals, generalizes effectively to unseen tasks without any fine-tuning, surpassing specialized state-of-the-art methods in several settings.
\makeatletter
\renewcommand\paragraph{\@startsection{paragraph}{4}{\z@}%
  {-12dd plus-4dd minus-4dd}%
  {-0.5em}%
  {\normalfont\normalsize\bfseries}} 
\makeatletter
\paragraph{Acknowledgments.}
This research is conducted within the ORCHID project, which receives funding from the French National Research Agency (ANR) (ANR-22-CE45-0029-01). For the purpose of open access, the authors have applied a CC BY public copyright license to any Author Accepted Manuscript (AAM) version arising from this submission. We acknowledge the financial support provided by PEPR Sharp (ANR-23-PEIA-0008, ANR, FRANCE 2030).

\paragraph{Disclosure of Interests.}
The authors have no competing interests to declare that are relevant to the content of this article.

%
%
%
\bibliographystyle{splncs04}
\bibliography{mybibliography}
%





\end{document}

%% file: settings/packages.tex
\usepackage{url}            
\usepackage{xspace}         
\usepackage{bm}

\usepackage{fancyhdr}       

\usepackage{etoolbox}

\usepackage{appendix}
\usepackage{multirow}
\usepackage{threeparttable}
\usepackage{graphicx}   
\usepackage{graphics}   
\usepackage{wrapfig}    
\usepackage{caption}    

\usepackage{tikz}     
\usetikzlibrary{shapes.geometric, arrows}
\usepackage[framemethod=TikZ]{mdframed}

\usepackage{xifthen}
\usepackage{xargs}

\usepackage[maxlevel=3]{csquotes} 
\usepackage{setspace}             
\usepackage{siunitx} 
\usepackage{amsmath}        
\usepackage{mathrsfs}       
\usepackage{amssymb}        
\usepackage{amsfonts}       
\usepackage{mathtools}      
\usepackage{dsfont}         
\usepackage{stmaryrd}       
\usepackage{esvect}         
\usepackage{systeme}        

\usepackage{tcolorbox}
\usepackage{enumitem}

\usepackage{booktabs}   
\usepackage{pifont}     
\usepackage{makecell}

\usepackage{hyperref}
\usepackage[capitalize]{cleveref}
\crefname{section}{Sec.}{Secs.}
\Crefname{section}{Section}{Sections}
\Crefname{table}{Table}{Tables}
\crefname{table}{Tab.}{Tabs.}

\usepackage[table,dvipsnames,svgnames]{xcolor}

%% file: settings/commandes_math.tex
\newcommand{\bracks}[1]{\left\lbrack #1 \right\rbrack} 
\newcommand{\pars}[1]{\left( #1 \right)}               

\newcommandx{\proba}[3][1, 3=0]{
\ifthenelse{\isempty{#2}}{\mathbb{P}_{#1}}{%
\ifthenelse{\equal{#3}{0}}{\mathbb{P}_{#1}\pars{#2}}{\mathbb{P}_{#1}\pars{#2 \middle| #3}}}%
}
\newcommandx{\esp}[3][1, 3=0]{
\ifthenelse{\isempty{#2}}{\mathbb{E}_{#1}}{%
\ifthenelse{\equal{#3}{0}}{\mathbb{E}_{#1}\bracks{#2}}{\mathbb{E}_{#1}\bracks{#2 \middle| #3}}}%
}
\newcommandx{\var}[3][1, 3=0]{
\ifthenelse{\isempty{#2}}{\operatorname{Var}_{#1}}{%
\ifthenelse{\equal{#3}{0}}{\operatorname{Var}_{#1}\pars{#2}}{\operatorname{Var}_{#1}\pars{#2 \middle| #3}}}%
}
\newcommandx{\cov}[4][1, 4=0]{
\ifthenelse{\isempty{#2} \AND \isempty{#3}}{\operatorname{Cov}_{#1}}{%
\ifthenelse{\equal{#4}{0}}{\operatorname{Cov}_{#1}\pars{#2, #3}}{\operatorname{Cov}_{#1}\pars{#2, #3 \middle| #4}}}%
}
\newcommandx{\corr}[3][1]{\ifthenelse{\isempty{#2} \AND \isempty{#3}}{\operatorname{Corr}_{#1}}{\operatorname{Corr}_{#1}\pars{#2, #3}}}

\newcommandx{\Norm}[3][1]{\mathcal{N}_{#1}\pars{#2, #3}} 

\newcommandx{\egalloi}{\overset{\mathrm{\mathcal{L}oi}}{=}} 
\newcommandx{\egalprob}{\overset{\mathbb{P}}{=}}            
\newcommandx{\egaltxt}[1]{\overset{\mathrm{#1}}{=}}         
\newcommandx{\simiid}{\overset{\mathrm{iid}}{\sim}}         

\newcommandx{\maxx}[1]{\underset{#1}{\max}}
\newcommandx{\minn}[1]{\underset{#1}{\min}}
\newcommandx{\argmax}[1][1]{\underset{#1}{\arg\max}}
\newcommandx{\argmin}[1][1]{\underset{#1}{\arg\min}}

\newcommandx{\integ}[4][2]{\int_{#1}^{#2} #3 \, \mathrm{d} #4} 

\newcommandx{\surf}[2][2=m]{\numprint{#1}\,#2\textsuperscript{2}}
\newcommandx{\vol}[2][2=m]{\numprint{#1}\,#2\textsuperscript{3}}

\newcommand*\fonction[5]{
#1 \colon \left\{\begin{alignedat}{2}  &#2 &\: &\to      #3\\
                                &#4 &   &\mapsto  #5
\end{alignedat} \right. \kern-\nulldelimiterspace} 


%% file: settings/custom.tex
\newcommand{\rot}[1]{\rotatebox{90}{#1}}

%% file: mybibliography.bib
@inproceedings{cai2025explore,
  title={Explore the Time Series Forecasting Potential of TabPFN Leveraging the Intrinsic Periodicity of Data},
  author={Cai, Sibo and Sun, Xi and Zhong, Hui},
  booktitle={1st ICML Workshop on Foundation Models for Structured Data},
  year={2025}
}

@article{hoo2025tables,
  title={From Tables to Time: Extending TabPFN-v2 to Time Series Forecasting},
  author={Hoo, Shi Bin and M{\"u}ller, Samuel and Salinas, David and Hutter, Frank},
  journal={arXiv preprint arXiv:2501.02945},
  year={2025}
}

@inproceedings{hoo2024tabular,
  title={The tabular foundation model tabpfn outperforms specialized time series forecasting models based on simple features},
  author={Hoo, Shi Bin and M{\"u}ller, Samuel and Salinas, David and Hutter, Frank},
  booktitle={NeurIPS workshop on time series in the age of large models},
  year={2024}
}

@inproceedings{
hollmann2023tabpfn,
title={Tab{PFN}: A Transformer That Solves Small Tabular Classification Problems in a Second},
author={Noah Hollmann and Samuel M{\"u}ller and Katharina Eggensperger and Frank Hutter},
booktitle={The Eleventh International Conference on Learning Representations },
year={2023},
url={https://openreview.net/forum?id=cp5PvcI6w8_}
}

@article{SU2024127063,
  title = {{{RoFormer}}: {{Enhanced}} Transformer with Rotary Position Embedding},
  author = {Su, Jianlin and Ahmed, Murtadha and Lu, Yu and Pan, Shengfeng and Bo, Wen and Liu, Yunfeng},
  year = 2024,
  journal = {Neurocomputing},
  volume = {568},
  pages = {127063},
  issn = {0925-2312},
  doi = {10.1016/j.neucom.2023.127063}
}

@inproceedings{
li2025mira,
title={{MIRA}: Medical Time Series Foundation Model for Real-World Health Data},
author={Hao Li and Bowen Deng and Chang Xu and ZhiYuan Feng and Viktor Schlegel and Yu-Hao Huang and Yizheng Sun and Jingyuan Sun and Kailai Yang and Yiyao Yu and Jiang Bian},
booktitle={The Thirty-ninth Annual Conference on Neural Information Processing Systems},
year={2025},
url={https://openreview.net/forum?id=Auy2DmlJBO}
}

@inproceedings{mullertransformers,
  title={Transformers Can Do Bayesian Inference},
  author={M{\"u}ller, Samuel and Hollmann, Noah and Arango, Sebastian Pineda and Grabocka, Josif and Hutter, Frank},
  year={2021},
  booktitle={International Conference on Learning Representations}
}

@InProceedings{He_2020_CVPR,
author = {He, Kaiming and Fan, Haoqi and Wu, Yuxin and Xie, Saining and Girshick, Ross},
title = {Momentum Contrast for Unsupervised Visual Representation Learning},
booktitle = {Proceedings of the IEEE/CVF Conference on Computer Vision and Pattern Recognition (CVPR)},
month = {June},
year = {2020}
}

@INPROCEEDINGS{7405423,
  author={López, S. and Suarez, G. and Jungreis, D. and Obeid, I. and Picone, J.},
  booktitle={2015 IEEE Signal Processing in Medicine and Biology Symposium (SPMB)}, 
  title={Automated identification of abnormal adult EEGs}, 
  year={2015},
  volume={},
  number={},
  pages={1-5},
  doi={10.1109/SPMB.2015.7405423}}

@article{wagnerPTBXLLargePublicly2020,
  title = {{{PTB-XL}}, a Large Publicly Available Electrocardiography Dataset},
  author = {Wagner, Patrick and Strodthoff, Nils and Bousseljot, Ralf-Dieter and Kreiseler, Dieter and Lunze, Fatima I. and Samek, Wojciech and Schaeffter, Tobias},
  year = 2020,
  month = may,
  journal = {Scientific Data},
  volume = {7},
  number = {1},
  pages = {154},
  issn = {2052-4463},
  doi = {10.1038/s41597-020-0495-6}
}

@article{hylandEarlyPredictionCirculatory2020,
  title = {Early Prediction of Circulatory Failure in the Intensive Care Unit Using Machine Learning},
  author = {Hyland, Stephanie L. and Faltys, Martin and H{\"u}ser, Matthias and Lyu, Xinrui and Gumbsch, Thomas and Esteban, Crist{\'o}bal and Bock, Christian and Horn, Max and Moor, Michael and Rieck, Bastian and Zimmermann, Marc and Bodenham, Dean and Borgwardt, Karsten and R{\"a}tsch, Gunnar and Merz, Tobias M.},
  year = 2020,
  month = mar,
  journal = {Nature Medicine},
  volume = {26},
  number = {3},
  pages = {364--373},
  publisher = {Nature Publishing Group},
  issn = {1546-170X},
  doi = {10.1038/s41591-020-0789-4},
  urldate = {2026-02-17},
  copyright = {2020 The Author(s), under exclusive licence to Springer Nature America, Inc.},
  langid = {english}
}

@article{pollardEICUCollaborativeResearch2018,
  title = {The {{eICU Collaborative Research Database}}, a Freely Available Multi-Center Database for Critical Care Research},
  author = {Pollard, Tom J. and Johnson, Alistair E. W. and Raffa, Jesse D. and Celi, Leo A. and Mark, Roger G. and Badawi, Omar},
  year = 2018,
  month = sep,
  journal = {Scientific Data},
  volume = {5},
  number = {1},
  pages = {180178},
  publisher = {Nature Publishing Group},
  issn = {2052-4463},
  doi = {10.1038/sdata.2018.178},
  urldate = {2026-01-22},
  copyright = {2018 The Author(s)},
  langid = {english}
}

@article{Andrzejak2001IndicationsON,
  title={Indications of nonlinear deterministic and finite-dimensional structures in time series of brain electrical activity: dependence on recording region and brain state.},
  author={Ralph G. Andrzejak and Klaus Lehnertz and Florian Mormann and Christoph Rieke and Peter David and Christian Erich Elger},
  journal={Physical review. E, Statistical, nonlinear, and soft matter physics},
  year={2001},
  volume={64 6 Pt 1},
  pages={
          061907
        },
  url={https://api.semanticscholar.org/CorpusID:8582357}
}

@article{Bagnall2018TheUM,
  title={The UEA multivariate time series classification archive, 2018},
  author={A. Bagnall and Hoang Anh Dau and Jason Lines and Michael Flynn and James Large and Aaron George Bostrom and Paul Southam and Eamonn J. Keogh},
  journal={ArXiv},
  year={2018},
  volume={abs/1811.00075},
  url={https://api.semanticscholar.org/CorpusID:53166683}
}

@article{Chen2015AGF,
  title={A general framework for never-ending learning from time series streams},
  author={Yanping Chen and Yuan Hao and Thanawin Rakthanmanon and Jesin Zakaria and Bing Hu and Eamonn J. Keogh},
  journal={Data Mining and Knowledge Discovery},
  year={2015},
  volume={29},
  pages={1622-1664},
  url={https://api.semanticscholar.org/CorpusID:12823839}
}

@article{Ng2018AnOA,
  title={An Open Access Database for Evaluating the Algorithms of Electrocardiogram Rhythm and Morphology Abnormality Detection},
  author={Eddie Y. K. Ng and Feifei Liu and Chengyu Liu and Lina Zhao and X. Zhang and Xiaoling Wu and Xiaoyan Xu and Yulin Liu and Caiyun Ma and Shoushui Wei and Zhiqiang He and Jianqing Li},
  journal={Journal of Medical Imaging and Health Informatics},
  year={2018},
  url={https://api.semanticscholar.org/CorpusID:70024401}
}

@inproceedings{10.1145/2939672.2939785,
author = {Chen, Tianqi and Guestrin, Carlos},
title = {XGBoost: A Scalable Tree Boosting System},
year = {2016},
isbn = {9781450342322},
publisher = {Association for Computing Machinery},
address = {New York, NY, USA},
url = {https://doi.org/10.1145/2939672.2939785},
doi = {10.1145/2939672.2939785},
booktitle = {Proceedings of the 22nd ACM SIGKDD International Conference on Knowledge Discovery and Data Mining},
pages = {785–794},
numpages = {10},
location = {San Francisco, California, USA},
series = {KDD '16}
}

@article{BaiTCN2018,
	author    = {Shaojie Bai and J. Zico Kolter and Vladlen Koltun},
	title     = {An Empirical Evaluation of Generic Convolutional and Recurrent Networks for Sequence Modeling},
	journal   = {arXiv:1803.01271},
	year      = {2018},
}

@inproceedings{dempster_etal_2021,
  author    = {Dempster, Angus and Schmidt, Daniel F and Webb, Geoffrey I},
  title     = {{MiniRocket}: A Very Fast (Almost) Deterministic Transform for Time Series Classification},
  booktitle = {Proceedings of the 27th ACM SIGKDD Conference on Knowledge Discovery and Data Mining},
  publisher = {ACM},
  address   = {New York},
  year      = {2021},
  pages     = {248--257}
}

@inproceedings{
jiang2024large,
title={Large Brain Model for Learning Generic Representations with Tremendous {EEG} Data in {BCI}},
author={Wei-Bang Jiang and Li-Ming Zhao and Bao-Liang Lu},
booktitle={The Twelfth International Conference on Learning Representations},
year={2024},
url={https://openreview.net/forum?id=QzTpTRVtrP}
}
